\documentclass[sigconf]{acmart}

\copyrightyear{2024} 
\acmYear{2024} 
\setcopyright{acmlicensed}\acmConference[ICCAD '24]{IEEE/ACM International
 Conference on Computer-Aided Design}{October 27--31, 2024}{New York, NY,
 USA}
 \acmBooktitle{IEEE/ACM International Conference on Computer-Aided Design
 (ICCAD '24), October 27--31, 2024, New York, NY, USA}
 \acmDOI{10.1145/3676536.3676747}
 \acmISBN{979-8-4007-1077-3/24/10}

\usepackage{multirow}
\usepackage{bbding}
\usepackage{bm}
\usepackage{pifont}
\usepackage{amsmath}
\usepackage{xcolor}
\usepackage{subfig}

\usepackage{algpseudocode}
\usepackage{algorithm}

\newcommand{\method}{MCUBERT}
\newcommand{\Emb}{\mathrm{Emb}}
\newcommand{\argmax}{\mathrm{argmax}}
\newcommand{\Diag}{\mathrm{Diag}}

\begin{document}

\title{\method: Memory-Efficient BERT Inference on Commodity Microcontrollers}


\author{Zebin Yang$^{1,2}$\footnotemark[1], Renze Chen$^{3}$\footnotemark[1], Taiqiang Wu$^{4}$, Ngai Wong$^{4}$, Yun Liang$^{2,6}$, Runsheng Wang$^{2,5,6}$, }
\author{Ru Huang$^{2,5,6}$, Meng Li$^{1,2,6}$\footnotemark[2]}

\affiliation{%
  \institution{$^1$Institute for Artificial Intelligence, Peking University, Beijing, China}
  \country{}
}
\affiliation{%
  \institution{$^2$School of Integrated Circuits, Peking University, Beijing, China}
  \country{}
}
\affiliation{%
  \institution{$^3$School of Computer Science, Peking University, Beijing, China}
  \country{}
}
\affiliation{%
  \institution{$^4$ The University of Hong Kong, Hong Kong, China}
  \country{}
}
\affiliation{%
  \institution{$^5$Institute of Electronic Design Automation, Peking University, Wuxi, China}
  \country{}
}
\affiliation{%
  \institution{$^6$Beijing Advanced Innovation Center for Integrated Circuits, Beijing, China}
  \country{}
}



\begin{abstract}

    In this paper, we propose \method~to enable language models like BERT on tiny microcontroller
    units (MCUs) through network and scheduling co-optimization.
    We observe the embedding table contributes to the major storage bottleneck for tiny BERT models.
    Hence, at the network level, we propose an MCU-aware two-stage neural architecture search algorithm
    based on clustered low-rank approximation for embedding compression.
    To reduce the inference memory requirements,
    we further propose a novel fine-grained MCU-friendly scheduling strategy.
    Through careful computation tiling and re-ordering as well as kernel design, 
    we drastically increase the input sequence lengths supported on MCUs without any latency or accuracy penalty.
    \method~reduces the parameter size of BERT-tiny and BERT-mini by 5.7$\times$
    and 3.0$\times$ and the execution memory
    by 3.5$\times$ and 4.3$\times$, respectively.
    \method~also achieves 1.5$\times$ latency reduction.
    For the first time, \method~enables lightweight BERT models on commodity
    MCUs and processing more than 512 tokens with less than 256KB of memory.

\end{abstract}
\maketitle
\pagestyle{plain} 

\renewcommand{\thefootnote}{\fnsymbol{footnote}}
\footnotetext[1]{Equal contribution} 
\footnotetext[2]{Corresponding author, meng.li@pku.edu.cn} 

\section{Introduction}
\label{sec:intro}

IoT devices based on microcontroller units (MCUs) are ubiquitous, enabling
a wide range of speech and language applications on the edge, including
voice assistant \cite{xu2019bert,chung2021w2v}, real-time translation \cite{zhu2020incorporating,yang2020towards}, 
smart home \cite{shin2019effective}, etc.
Language models (LMs), e.g., BERT \cite{devlin2018bert}, are fundamental to these applications.
While cloud off-loading is heavily employed for LM processing, it suffers from high latency
overhead, privacy concerns, and a heavy reliance on WiFi or cellular 
networks \cite{banbury2021micronets,fedorov2022udc}.
Hence, there is a growing demand for deploying BERT-like LMs on MCUs.

\begin{figure}[!tb]
    \centering
    \includegraphics[width=1.0\linewidth]{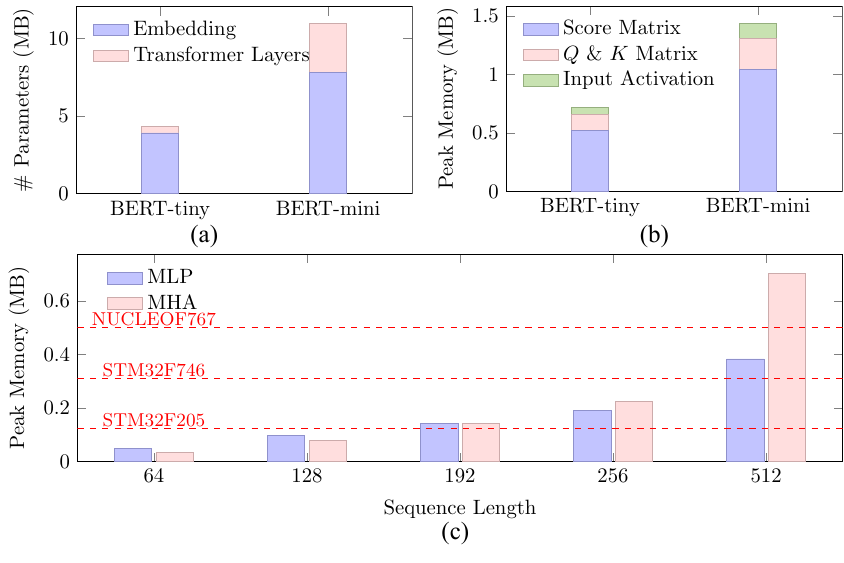}
    \caption{Enabling BERT on MCUs faces memory challenges: (a) the Flash storage 
    limits the model size; (b) the SRAM memory limits the peak execution memory;
    (c) for long sequence lengths, memory requirements of both MHA and multi-layer
    perceptron (MLP) become bottleneck.}
    \label{fig:intro:mem_challenge}
    \vspace{-15pt}
\end{figure}

Though appealing, enabling BERT on MCUs is very challenging. On one
hand, MCUs only have a very simple memory hierarchy with highly constrained
memory budgets \cite{banbury2021micronets,burrello2021microcontroller}.
For example, a state-of-the-art (SOTA) ARM Cortex-M7 MCU only has
320 KB static random-access memory (SRAM) to store intermediate data, e.g., activation, and 1 MB
Flash to store program and weights, directly
limiting the peak execution memory and the total parameter size
of LMs \cite{lin2020mcunet,banbury2021micronets,burrello2021microcontroller,lin2021mcunetv2}.
As shown in Figure~\ref{fig:intro:mem_challenge},
even with 8-bit quantization,
the BERT-tiny model \cite{bhargava2021generalization,DBLP:journals/corr/abs-1908-08962}
exceeds the SRAM and Flash capacity by more than 2.3$\times$ and
4.3$\times$, respectively. 

On the other hand, the computation graph of a Transformer block in BERT is more complex
when compared with convolutional neural networks (CNNs):
each multi-head attention (MHA) block not only comprises more MCU-unfriendly 
tensor layout transformation operators, e.g., reshape, transpose, etc,
but also organizes them with linear operators in a more complex topology.
Naively executing these operators can result in significant memory consumption and poor performance 
\cite{ivanov2021data,burrello2021microcontroller}.

Existing works, however, cannot resolve these challenges. Most SOTA BERT 
designs \cite{jiao2019tinybert,lan2019albert,sanh2019distilbert}
and network optimization designs
\cite{xu2021bert,zhang2022autodistill}
focus on large BERT models on GPUs or smartphones. As the memory and
storage of these platforms are much more abundant than MCUs, these methods
usually focus on optimizing the Transformer layers for better latency.
Another class of works \cite{fedorov2019sparse,lin2020mcunet,lin2021mcunetv2,banbury2021micronets}
develop MCU-friendly CNNs by jointly optimizing the
network architecture and scheduling, which, however, cannot be directly applied to BERT.

In this paper, we propose \method, a network and scheduling
co-optimization framework to enable BERT on commodity MCUs for the first time. 
We observe for small BERT models, e.g., BERT-tiny, the embedding table accounts
for the major storage bottleneck and thus,
propose to leverage the MCU-friendly clustered low-rank approximation for embedding compression.
We further propose an MCU-aware two-stage differentiable neural architecture search (NAS) algorithm to improve the
accuracy of compressed models.
To reduce the execution memory usage and latency, we observe the execution of 
a Transformer block can
be carefully tiled without accuracy penalty and develop an MCU-friendly
fine-grained scheduling algorithm.
Our contributions can be summarized as follows:
\begin{itemize}
    \item We propose an MCU-aware two-stage NAS algorithm based on clustered
      low-rank approximation for embedding compression to overcome the storage
      bottleneck.
    \item To reduce the execution peak memory and latency, we propose
      MCU-friendly scheduling optimization for the Transformer block.
    \item \method~can reduce the model size by 5.7$\times$ and 3.0$\times$ as well as
    the peak memory by 3.5$\times$ and 4.3$\times$ for BERT-tiny
    and BERT-mini, respectively, enabling to process more than 512 tokens
    simultaneously with less than 256KB SRAM. \method~also achieves 1.5$\times$ and 1.3$\times$
    latency reduction compared to SOTA inference engines, CMSIS-NN and \cite{burrello2021microcontroller},
    respectively.
\end{itemize}

\section{Related Works}
\label{sec:prelim}

\begin{table}[!tb]
\centering
\caption{Comparison with prior-art methods (Opt represents Optimization).}
\label{tab:previous_work}
\scalebox{0.8}{
\begin{tabular}{c|cccccc}
\toprule
 \multirow{2}{*}{Method}&\multirow{2}{*}{Platform}&\multirow{2}{*}{Model}&\multirow{2}{*}{Network Opt}& \multicolumn{3}{c}{Scheduling Opt} \\
 & & &&MLP &MHA &Kernel\\
\midrule
\cite{banbury2021micronets}                              & MCU & CNN         & Conv       & -             &  -  &  \ding{55}       \\
\cite{lin2020mcunet,lin2021mcunetv2}                     & MCU & CNN         & Conv       & -  & - & \Checkmark\\
  \cite{liang2023mcuformer}            & MCU & ViT    & MLP       & \Checkmark   &  \ding{55}   &\Checkmark        \\
\midrule
\cite{sanh2019distilbert}                                & GPU & BERT
  & Linear     & \ding{55}    &\ding{55} &\ding{55}    \\
\cite{dao2022flashattention}                             & GPU & BERT        & -        &  \ding{55} &\Checkmark &\Checkmark\\
\midrule
Ours                                                     & MCU & BERT        & Embedding       &  \Checkmark & \Checkmark &\Checkmark  \\ 
\bottomrule
\end{tabular}
}
\vspace{-5pt}
\end{table}

\subsection{Model Deployment on MCUs.}
\label{app:model deployment on mcus}
There are two main approaches to deploy models on MCUs, 
interpretation \cite{chen2018tvm,david2021tensorflow} and code generation \cite{lin2020mcunet,lin2021mcunetv2,burrello2021microcontroller}. 
The interpretation-based methods embed an on-device runtime interpreter to support flexible model deployment
but require extra memory to store meta-information and extra time for runtime interpretation. 
Code-generation-based methods directly compile the given model into target code to save memory and reduce inference latency. 
\method~uses a code generation method that is more specialized for our target model and device,
leading to lower latency and lower memory usage.

\subsection{Network efficiency optimization.}
Network efficiency is very important for the overall performance of deep learning systems and has been widely studied.
We focus on reviewing the model optimizations targeting at MCUs and for LMs.

Deep learning models need to meet the tight storage and memory constraints to run on MCUs.
Previous works propose network and scheduling optimization as summarized in Table \ref{tab:previous_work}. 
\cite{rusci2020memory,banbury2021micronets,lin2020mcunet} compress CNNs with NAS to meet the storage constraints.
\cite{burrello2021microcontroller} deploys small Transformer models which already satisfy the memory constraints of MCUs.
\cite{liang2023mcuformer} deploys vision transformer (ViT) on MCUs mainly by compressing the MLP layers and searching the token numbers. However, there are new challenges for deploying BERT on MCUs. First, compared to tiny CNNs and small Transformer models, BERT models have more parameters. Second, the memory bottleneck of BERT inference mainly lies in the MHA block when sequence length is long, which is shown in Figure \ref{fig:intro:mem_challenge}(c). The computation of MHA block is more complex compared with CNN blocks and MLP layers, bringing new challenges for scheduling optimization. We propose MCUBERT to first deploy BERT, the most representative encoder transformer model, on MCUs.

Though no existing works deploy BERT models on MCUs, there are network optimization algorithms proposed
for LMs on other platforms, e.g., GPUs,
such as pruning \cite{shi2021sparsebert, cui2019fine, zafrir2021prune, campos2022sparse}, quantization \cite{bai2020binarybert,zhang2020ternarybert,shen2020q, qin2022bibert}, low-rank factorization \cite{baevski2018adaptive, chen2018groupreduce, hsu2022language, lan2019albert, lioutas2019distilled}, etc. 
As MCUs do not natively support low-bit quantization or sparse computation, low-rank factorization is usually more MCU-friendly.
Existing works such as Distilled Embedding \cite{lioutas2019distilled} and Albert \cite{lan2019albert} 
leverage low-rank factorization based on singular value decomposition (SVD) for embedding compression and prune
singular values with small magnitudes.
Adaptive embedding \cite{baevski2018adaptive} compresses the embedding table with clustered low-rank approximation:
it divides the tokens in an embedding table into clusters first and applies low-rank
approximation with different ratios to each cluster based on the cluster importance.
Adaptive embedding empirically leverages token frequency as the proxy metric for its importance, which leads to high accuracy degradation.

There are also scheduling optimization proposed for Transformer models on GPUs.
FlashAttention \cite{dao2022flashattention} only computes the attention score tensor partially each time and repetitively
update the accumulation of the partial sum to drastically reduce the execution memory. 
There are also previous works like \cite{zhai2023bytetransformer,fang2021turbotransformers,rasley2020deepspeed} 
that use kernel fusion to reduce memory usage and accelerate inference.
However, these methods usually do not consider quantization and targets at both training and inference.
Our scheduling optimization is inspired by these methods but is more MCU-friendly.

As shown in Table~\ref{tab:previous_work}, our proposed \method~first deploys transformer model BERT on MCUs. \method~compresses the embedding table, which account for most of the parameters and can't be stored in MCU.
Besides MLP block, We also carefully optimize the MHA block, which comprises more MCU-unfriendly operators and can also be the memory bottleneck of BERT inference.
Compared with GPU-based methods, \method~conduct network and scheduling optimization in a more MCU-friendly mode.

\section{\method: MCU-friendly Network/Scheduling Co-Optimization}
\label{sec:Method}

\begin{table}[!tb]
  \caption{Notations used in the paper.}
\label{tab:notations}
\centering
\scalebox{0.78}{
\begin{tabular}{c|c}
\toprule
Notations           & Meanings \\
\midrule
$v$                 & Vocabulary size \\
$s, h, d$           & Sequence length, \# heads, and embedding dimension        \\
$c$                 & \# clusters                \\
$i,j,l$             & Loop variables for clusters, tokens, and singular values \\
$t$                 & Sequence length each tile     \\
$U, V$              & Unitary matrices generated by SVD         \\
$U_i, V_i$              & Embedding table and linear projection for the $i_{th}$
 cluster        \\
$U_{i,j}$              & Embedding vector for the $j_{th}$
 token in $i_{th}$ cluster        \\
$\Sigma$            & Vector of singular values         \\
$\alpha, \beta$     & NAS parameters for embedding compression \\
$\alpha_{j,i}$  &   NAS parameters for $j_{th}$ token in $i_{th}$ cluster in first stage NAS \\
$\beta_{i,l}$     & NAS parameters for $l_{th}$
 singular value in $i_{th}$
 cluster in second stage NAS \\
$\beta_i^*$     & 	Threshold for NAS parameters in $i_{th}$
 cluster in second stage NAS \\
$M,N,K$& Loop ranges of a matrix multiplication \\
$m,n,k$             & Loop variables of a matrix multiplication \\
\bottomrule 
\end{tabular}
}
\end{table}

\subsection{Motivations and Overview}

We now discuss our observations on key challenges that prevent running
BERTs directly on MCUs and introduce the overall flow of \method.
The notations used in the section is summarized in Table~\ref{tab:notations}.



\paragraph{\textbf{Observation 1: the tight MCU Flash storage limits model size and forces to use 
small BERT models, for which the embedding table becomes the major bottleneck.}}
To satisfy the Flash storage constraints of MCUs, which is often less than 2 MB
\cite{lin2020mcunet,banbury2021micronets}, we are forced to consider only small BERT
models, e.g., BERT-tiny and BERT-mini \cite{devlin2018bert},
which have lower embedding dimensions and fewer Transformer layers.
However, there still exists a 4.3$\times$ gap between the model size and the MCU Flash,
even for BERT-tiny with 8-bit quantization. As shown in Figure~\ref{fig:intro:mem_challenge}(a),
the embedding table contributes to more than 90\% of the parameters of
BERT-tiny and becomes the major bottleneck. Hence, embedding compression is required
to enable BERT on MCUs.


\paragraph{\textbf{Observation 2: the MCU SRAM size limits the execution peak memory
of both MHA and MLP, especially for long sequence lengths.}}
During BERT inference, all the activations need to be stored in the MCU SRAM.
Although Flash paging and re-materialization has been proposed \cite{patil2022poet}
to reduce the SRAM requirements,
the introduced Flash access and re-computation incur high power and latency overhead.
As shown in Figure~\ref{fig:intro:mem_challenge}(c), with the increase of
the sequence length, the peak memory of both MHA and MLP increases significantly and
quickly exceed the MCU SRAM limit. Depending on the sequence length, both MLP and MHA
can be the memory bottleneck. Meanwhile, as shown in Figure~\ref{fig:intro:mem_challenge}(b),
for a long input sequence, the score matrix in MHA incurs highest memory consumption.
Therefore, both MLP and MHA, especially the score matrix, needs to be optimized to enable
processing long sequences.


\paragraph{\textbf{Observation 3: the naive design of computation kernels introduces non-negligible latency overhead.}}
Both MHA and MLP involve many linear/batched matrix multiplication operators with large shapes.
MHA further complicates the inference with extensive tensor shape transformation operators, e.g., reshape, transpose, etc.
Naive kernel implementation, e.g., CMSIS-NN, cannot well utilize the hardware resource and suffers from
over 30\% latency overhead. To improve the resource utilization, it is important to leverage the memory locality
and the instruction-level parallelism (ILP) of MCU. Memory access patterns can also be designed to fuse
the tensor shape transformation operators with the linear operators. All of these require dedicated kernel optimization.

\begin{figure}[!tb]
    \centering
    \includegraphics[width=\linewidth]{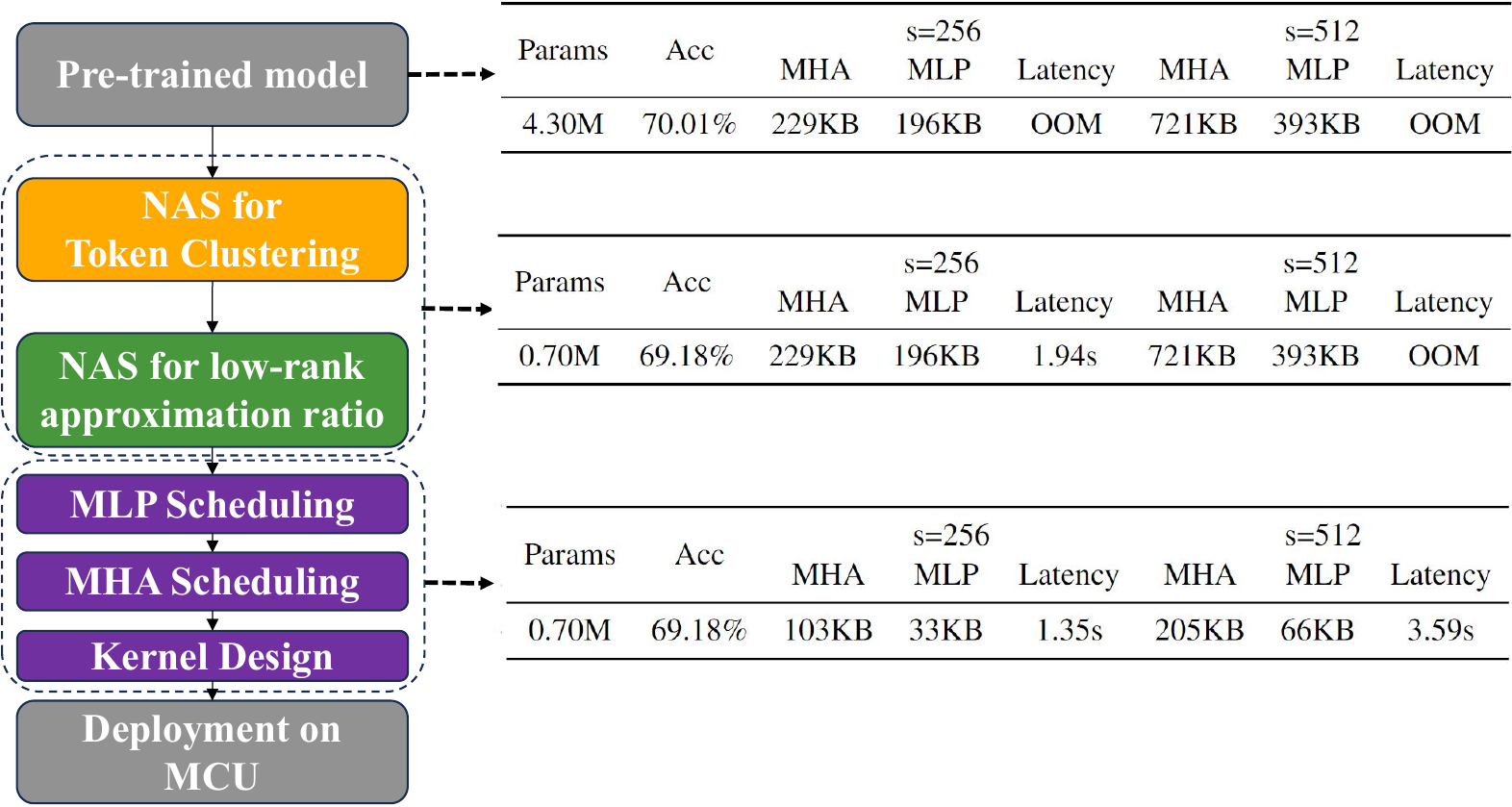}
    \caption{MCUBERT overview. (Params stands for parameters, Acc stands for MNLI accuracy, and OOM stands for out of memory.) }
    \label{pipeline}
    \vspace{-10pt}
\end{figure}

\paragraph{\textbf{\method~overview.}} Based on these observations, we propose \method, an MCU-friendly network/scheduling
co-optimization framework to enable BERT models on tiny devices. The overview of \method~is shown in Figure~\ref{pipeline}.
\method~leverages MCU-friendly clustered low-rank approximation for embedding compression and features a MCU-aware two-stage NAS
to explore the trade-off between accuracy and model sizes, enabling to reduce the model size 
and fit BERT models into the MCU Flash storage.
To enable processing long sequences, \method~proposes MCU-friendly scheduling optimization.
By tiling and re-ordering the computation as well as designing efficient kernels,
inference peak memory and latency can be reduced without accuracy penalty.
Such optimization enables to support more than 512 tokens on a small MCU with less than 256 KB SRAM.

Note while we focus on optimizing BERT-like encoder LMs for MCUs, our proposed techniques,
including embedding compression and scheduling optimization, can benefit GPT-like decoder LMs \cite{radford2019language}
on MCUs as well. Decoder LMs face more challenges on the storage and dynamic
shape induced by the KV cache, which we leave for future research.


\subsection{MCU-aware NAS for Embedding Compression}
\label{subsec:nas}

\paragraph{\textbf{MCU-aware NAS Formulation.}}
To reduce the parameter size and satisfy the tight Flash storage constraint, we propose 
embedding compression based on clustered low-rank approximation
following \cite{baevski2018adaptive}, which is more MCU-friendly as MCUs do not
natively support sub-8-bit quantization or sparse computation.
Following \cite{baevski2018adaptive}, we set the number of clusters to 4 and show its impact in ablation study.
Two important questions remain to be answered:
1) how to cluster the tokens and 2) how to determine the approximation ratio for each cluster.
We propose differentiable NAS to guide the compression directly with task loss.


\begin{figure}[!tb]
    \centering
    \includegraphics[width=\linewidth]{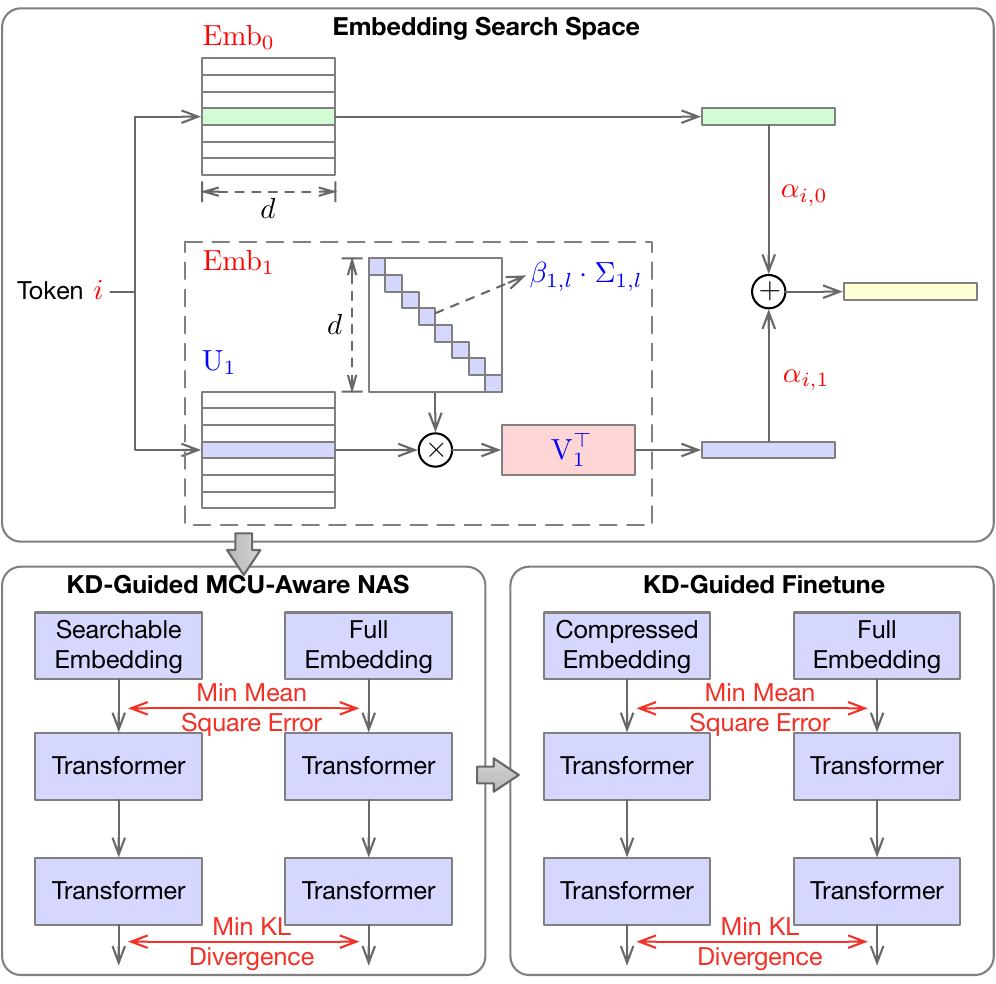}
    \caption{Our proposed NAS search formulation for embedding compression.}
    \label{fig:nas}
\end{figure}

Let $c$ denote the number of clusters and $\Emb_{i}$ denote the embedding table for $i$-th cluster,
where $0 \leq i \leq c - 1$. To enable NAS for token clustering, as shown in Figure~\ref{fig:nas},
we introduce trainable architecture parameters $\alpha$ ($0 \leq \alpha \leq 1$) for each token in the vocabulary.
Let $v$ denote the vocabulary size. Then, for the $j$-th token ($0 \leq j \leq v - 1$),
we define $\{\alpha_{j, 0}, \ldots, \alpha_{j, c-1}\}$,
where $\sum_{i=0}^{c-1} \alpha_{j, i} = 1$. Then, the embedding for the $j$-th token becomes:

\begin{equation}
\begin{aligned}
    \alpha_{j, 0} \Emb_{0}[j] + \sum_{i=1}^{c-1} \alpha_{j, i} \Emb_{i}[j],
\end{aligned}
\end{equation}

where $\Emb_{i}[j]$ denotes the embedding vector of $j$-th token from the $i$-th embedding table.

To leverage NAS for low-rank approximation, instead of directly searching the reduced
dimension for each embedding table, we apply singular value decomposition (SVD)
and search the singular values that can be pruned,
which enables to better inherit the pretrained weights and is crucial for BERT compression.
Specifically, we first decompose $\Emb_{i}$ with SVD as $U_i \Diag(\Sigma_i) V_i^{\top}$,
where $U_i$ and $V_i$ are unitary matrices, $\Sigma_i$ is the vector of singular values,
and $\Diag(\Sigma_i)$ represents the diagonal matrix with singular values in $\Sigma_i$ filled in the diagonal.
We also introduce architecture parameters $\beta$ ($0 \leq \beta \leq 1$) for each cluster. For the $i$-th cluster,
we define $\{\beta_{i, 0}, \ldots, \beta_{i, d-1}\}$, where $d$ is the
embedding dimension, 
to indicate the importance of each singular value. Then, we have
\begin{equation}
\begin{aligned}
    \Emb_{i} = U_i \Diag(\beta_{i, 0} \Sigma_{i, 0}, \ldots, \beta_{i, d-1} \Sigma_{i, d-1}) V_i^{\top}.
\end{aligned}
\end{equation}
Based on $\alpha$ and $\beta$, the embedding size for the $i$-th cluster can be approximated as
\begin{equation}
\label{equ:method:size_emb}
\begin{aligned}
    \mathrm{Size}(\Emb_i) = \sum_{l} \beta_{i, l} (d + \sum_{j} \alpha_{j, i}),
\end{aligned}
\end{equation}
where $\sum_{j} \alpha_{j, i}$ approximates the number of tokens for $i$-th cluster.
$\sum_{l} \beta_{i, l} \sum_{j} \alpha_{j, i}$ and $\sum_{l} \beta_{i, l} d$
represent the size of low-rank approximated $U_i$ and $V_i^\top$, respectively.
We introduce $\ell_1$-penalty of the embedding table size into the objective function 
to encourage a lower parameter size:
\begin{equation}
\begin{aligned}
    \min_{\alpha, \beta} \min_{w} \ell_{w, \alpha, \beta} + \lambda \sum_{i} \mathrm{Size}(\Emb_i).
\end{aligned}
\end{equation}
where $w$ denotes the model parameters and $\lambda$ is the hyperparameter to balance model size and accuracy.
$\alpha$, $\beta$, and $w$ are trained together. Upon convergence, for the $j$-th token, 
we determine its cluster by $\argmax_{i} \alpha_{j, i}$, while for the $i$-th cluster,
we set a threshold $\beta_i^*$ and only keep the $l$-th singular values with $\beta_{i, l} > \beta_{i}^*$.

\paragraph{\textbf{Improving NAS Convergence.}}
In practice, we find that searching token clustering and low-rank approximation ratio, i.e., $\alpha$ and $\beta$,
simultaneously makes the differentiable NAS training unstable and directly degrades the accuracy of the searched models.
As shown in Figure~\ref{fig:importance of two-stage}, the accuracy of the searched models is even lower compared to the baseline
adaptive embedding \cite{baevski2018adaptive}.
We hypothesize this is because of the large discrepancy of optimal low-rank approximation ratios for different
clusters. Because the token clustering keeps changing during NAS, some singular values in a certain cluster may be
incorrectly preserved due to tokens that eventually belong to a different cluster. These incorrectly preserved singular
values in turn may impact the token clustering, leading to sub-optimal results.

To encourage the NAS convergence, we decompose the search space
and propose a two-stage NAS strategy: in the first stage,
we fix different low-rank approximation ratios for each cluster,
and only search the token clustering, i.e., $\alpha$. The approximation ratios are
chosen following \cite{baevski2018adaptive}. Then, in the second stage, given the
fixed token clustering, we search the best approximation ratio for each cluster, i.e., $\beta$.
While we can iterate back to the first stage with updated approximation ratios,
we empirically find it is not necessary. We hypothesize this is because the token clustering reflects the token importance,
which should remain stable for different low-rank approximation ratios.
The algorithm details are shown in Algorithm \ref{alg:method:1}, \ref{alg:method:2}, and \ref{alg:method:3}.
We guide the NAS with knowledge distillation (KD) to improve convergence following \cite{sanh2019distilbert}.
We simply use the trained full model as the teacher.

\begin{algorithm}[!tb]
\caption{Searching for Token Clustering}
\label{alg:method:1}
\begin{algorithmic}[1]
\Require Embedding matrix $Emb$, vocabulary size $v$, \# clusters $c$, embedding dimension $d$, division value $div$, \# training epochs $epochs$
\Ensure Clustering results $cls$
\State $\alpha_{j,i} = \frac{1}{c} \quad \forall j, i$
\State $U_0 = Emb$ 
\Statex\Comment{The embedding table of first cluster is not factorized}
\For{$i = 1, \dots, c-1$}
    \State $U_i, V_i^T = \text{LowRankFactorization}(Emb, \frac{d}{div^i})$ 
    \Statex\Comment{Factorize the embedding table of other clusters except }
    \Statex\quad \ \ \quad \ \ the first cluster
\EndFor
\State Define the embedding of $j$-th token: $\alpha_{j,0}U_{0,j} + \sum_{i=1}^{c-1} \alpha_{j,i}U_{i,j}V_i^T$ 
 \Statex\Comment{$V_0$ is excluded as the embedding table of the first cluster is }
 \Statex\quad \ \ not factorized
\For{$z = 0, \dots, epochs-1$}
    \State Fix $\alpha$ and update weights $w$ by descending $\nabla_w \ell(w, \alpha)$
    \State Fix $w$ and update architecture parameters $\alpha$ by descending 
    \Statex\quad \ \ $\nabla_\alpha (\ell(w, \alpha) + \lambda \sum_i \text{Size}(Emb_i))$
\EndFor
\State $cls[j] = \arg \max_i (\alpha_{j,i}) \quad \forall j$ \\
\Return $cls$
\end{algorithmic}
\end{algorithm}

\begin{algorithm}[!tb]
\caption{LowRankFactorization}
\label{alg:method:2}
\begin{algorithmic}[1]
\Require  Matrix to be factorized $Matrix$; Factorization ratio $r$
\Ensure  Factorization Results $U$ and $V^T$
\State $U, \text{Diag}(\Sigma), V^T = \text{SVD}(Matrix)$ 
\Statex\Comment{$\text{Diag}(\Sigma)$ denotes the diagonal matrix filled with vector $\Sigma$}
\State $\Sigma = \Sigma[0 : r]$
\State $U = U[:, 0 : r] ~\text{Diag}(\Sigma^{1/2})$
\State $V^T = \text{Diag}(\Sigma^{1/2})~V^T[0 : r, :]$ \\
\Return $U, V^T$
\end{algorithmic}
\end{algorithm}

\begin{algorithm}
\caption{Searching for Low-Rank Approximation Ratio}
\label{alg:method:3}
\begin{algorithmic}[1]
\Require Embedding matrix $Emb$, vocabulary size $v$, \# clusters $c$, embedding dimension $d$, \# training epochs $epochs$, importance threshold $\beta^*$, searched token clustering $cls$
\Ensure Low-rank approximation ratios $ratios$
\State $\beta_{i,l} = \beta_i^* \quad \forall i,l$
\State Divide the $Emb$ into $c$ parts $Emb_0, \ldots, Emb_{c-1}$ according to token clustering $cls$
\State $U_0 = Emb_0$ 
\Statex\Comment{The embedding table of first cluster is not factorized}
\For{$i = 1, \ldots, c-1$}
    \State $U_i, \text{Diag}(\Sigma_{i,0}, \ldots, \Sigma_{i,d-1}), V_i^T = \text{SVD}(Emb_i)$ 
    \Statex\Comment{
    Factorize the embedding table of other clusters except the}
    \Statex \quad \ \ first cluster
\EndFor
\State Define embedding of $j$-th token in $i$-th cluster (for $i \geq 1$): $U_{i,j} \cdot \text{Diag}(\beta_{i,0}\Sigma_{i,0}, \ldots, \beta_{i,d-1}\Sigma_{i,d-1})V_i^T$
\State For the first cluster, the embedding of $j$-th token is $U_{0,j}$
\For{$z = 0, \ldots, epochs-1$}
    \State Fix $\beta$ and update weights $w$ by descending $\nabla_w \ell(w, \beta)$
    \State Fix $w$ and update architecture parameters $\beta$ by descending 
    \Statex \quad \ \ $\nabla_\beta(\ell(w, \beta) + \lambda \sum_i \text{Size}(Emb_i))$
\EndFor
\State Initialize $ratios[i] = 0 \quad \forall i$
\For{$i = 1, \ldots, c-1$}
    \For{$l = 0, \ldots, d-1$}
        \State $ratios[i] += 1$ \quad \textbf{if} $\beta_{i,l} > \beta_i^*$
        \Statex \Comment{ Only save singular values whose importance is larger}
        \Statex\quad \ \quad \  \ \ than threshold
    \EndFor
\EndFor \\
\Return $ratios$
\end{algorithmic}
\end{algorithm}

Our proposed two-stage NAS incurs low training cost as only a few training epochs (less than 2) are needed
for each stage. After the two-stage NAS, we quantize both the weights and activations
to 8 bits and fine-tune the compressed model with KD to get the final deployable model. 

\subsection{MCU-friendly Scheduling Optimization}
\label{sec:schedule_opt}


We now introduce our MCU-friendly scheduling optimization to reduce the inference memory and latency.
We optimize the scheduling of both MHA and MLP since and also optimize the kernel implementation for MHA for better efficiency.


\begin{figure}[!tb]
    \centering
    \includegraphics[width=0.9\linewidth]{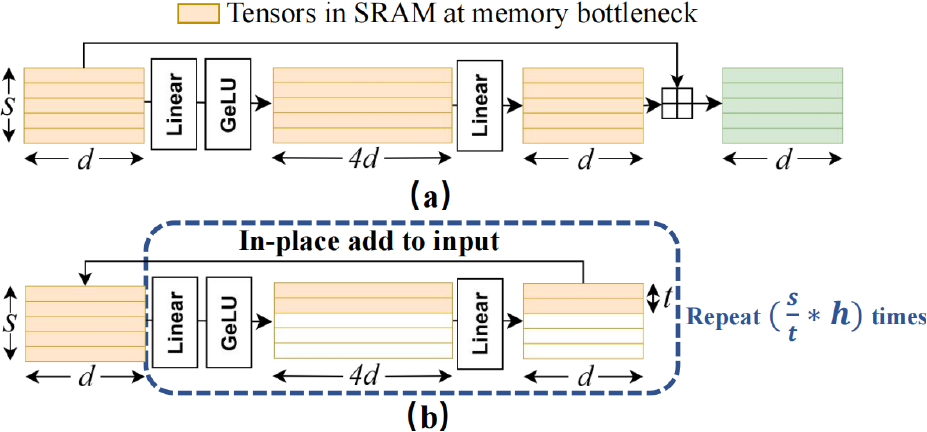}
    \caption{MLP scheduling to reduce peak memory. The tensor in yellow will be saved in SRAM at memory bottleneck. }
    \label{fig:schedule:mlp}
    \vspace{-10pt}
\end{figure}

\paragraph{\textbf{MLP Scheduling.}}
The memory bottleneck of MLP comes from the second linear layer. 
As shown in Figure~\ref{fig:schedule:mlp}(a), 3 tensors need to be stored,
whose sizes add up to $6 \cdot s \cdot d$.
We observe the computation of different tokens in an MLP is independent. This
enables us to divide the input activation into tiles along the token dimension.
Assume each tile has $t$ tokens. Then, we can re-order the computation to finish all the MLP computations
for one tile before moving to the next tile. Moreover, we can also perform an in-place
addition with the residual and directly overwrite the $t$ input tokens as shown in Figure~\ref{fig:schedule:mlp}(b). 
Thereby, we can reduce the execution memory to $s \cdot d + t \cdot 5d$. 
As $t$ is usually much smaller than $s$, 6$\times$ memory reduction can be achieved.
In practice, by analytically computing the relation between $t$ and the peak memory usage, we can directly choose $t$ based on the MCU SRAM size.
Hence, such scheduling optimization incurs negligible latency overhead on commodity MCUs.


\paragraph{\textbf{MHA Scheduling.}}
Unlike MLP, the computation of different tokens depends on each other,
making it hard to fully tile the MHA computation.
We observe the following optimization opportunities: 1) the computation of each head is independent,
and 2) the score tensor accounts for the major bottleneck, shown in Figure \ref{fig:intro:mem_challenge},
and its computation can be tiled along token dimension.

\begin{figure}[!tb]
    \centering
    \includegraphics[width=1.0\linewidth]{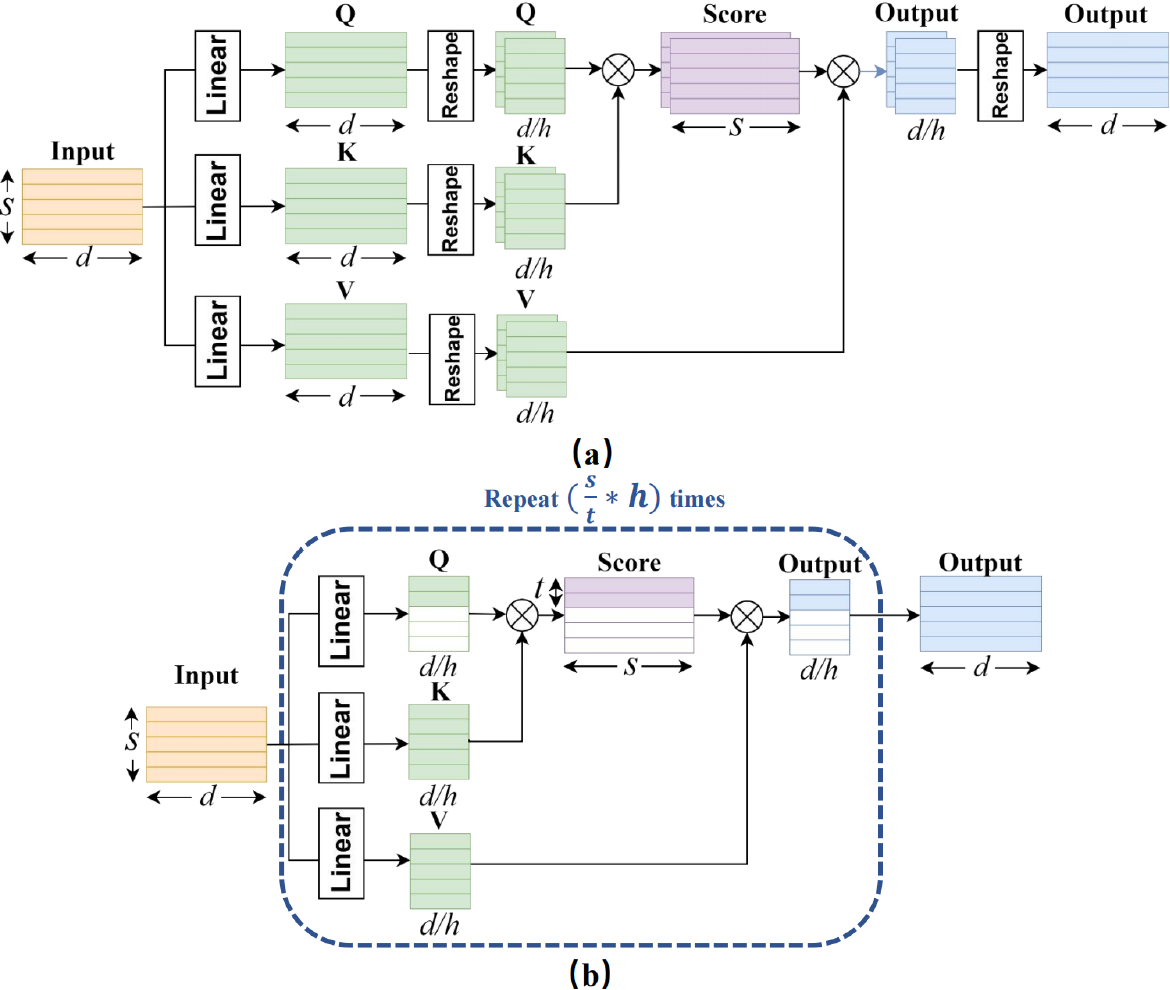}
    \caption{MHA scheduling to reduce the tensor transformation latency and peak memory.}
    \label{fig:schedule:mha}
\end{figure}

Based on the observation, we propose a new MHA scheduling in Figure~\ref{fig:schedule:mha}(b).
We first tile the computation along the head dimension, i.e., $h$, and then,
further tile the query tensor along the token dimension. This indicates
we compute the attention between $t$ tokens and all the $s$ tokens per head each time.
It enables us to reduce the memory of the score matrix,
breaking the quadratic increase of execution memory into a linear increase with $s$.
Again, we carefully choose $t$ considering both memory constraints and computation parallelism to minimize latency overhead.

Our proposed MHA scheduling shares similarities with FlashAttention \cite{dao2022flashattention}, which tiles the key and value matrix.
However, FlashAttention requires updating the output matrix repetitively.
Either the output matrix has to be stored in high precision,
leading to a higher peak execution memory, or quantization and de-quantization operations are
needed during each output accumulation step, bringing heavy computation pressure and accuracy loss.
Our method is more MCU-friendly and will not face such issues.


\paragraph{\textbf{Kernel Design Optimization.} }
The scheduling optimization reduces the peak memory for MLP and MHA.
To reduce latency, we also design optimized kernels for each tile of computation.

\begin{figure}[!tb]
    \centering
    \includegraphics[width=1.0\linewidth]{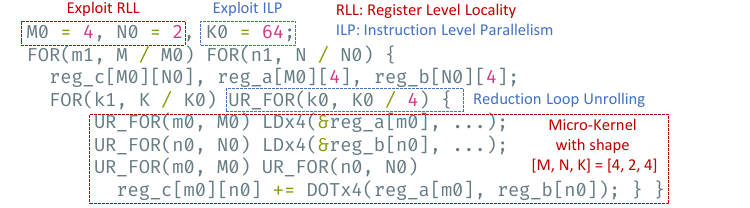}
    \caption{Matrix multiplication kernel design of \method. 
    \texttt{UR\_FOR} means fully unrolled for-loop.
    \texttt{LDx4} means loading 4 consecutive elements from SRAM to the register. \texttt{DOTx4} means computing dot-product of two 4-element vectors using SIMD instruction (e.g., \texttt{SMLAD} for ARM Cortex-M MCU). The quantization and writing-back operations are omitted here.}
    \label{fig:schedule:kernel}
\end{figure}

First, we apply a two-level loop blocking to better fit into the Register-SRAM hierarchy of MCU.
Figure~\ref{fig:schedule:kernel} illustrates the kernel design for each tile of the matrix multiplication/linear operator.
The innermost block is referred to as a micro-kernel,
which computes entirely on registers. Properly setting its shape helps exploit register-level locality, thus reducing inference latency. In our experiments, we set the micro-kernel shape to [M, N, K] = [4, 2, 4]. In comparison, CMSIS-NN~\cite{lai2018cmsis} employs a shape of [1, 2, 4], which exhibits lower locality and decreases performance.
Second, we unroll the reduction loop with a factor of 64 to harness the ILP of MCU and better utilize
the hardware instruction pipeline.
Moreover, by designing the memory access patterns for the linear operators,
we fuse all the tensor shape transformation operators and avoid dedicated memory accesses.
The kernel design optimization enables to fully leverage the hardware characteristics of MCUs
for more efficient BERT processing.

\section{Experiments}
\label{sec:Experiments}


\subsection{Experiment Setup}

\paragraph{\textbf{Dataset.}} We search and evaluate our models on the General
Language Understanding Evaluation (GLUE) benchmark \cite{wang2018glue},
which is a collection of text classification tasks. 
For most evaluations and comparisons, we leverage the Multi-Genre Natural Language Inference Corpus (MNLI) dataset,
which is the largest dataset in GLUE. 


\paragraph{\textbf{Searching setting.}}
We select lightweight BERT models, i.e., BERT-tiny and BERT-mini for our experiments, and the pre-trained models are adopted from \cite{bhargava2021generalization,DBLP:journals/corr/abs-1908-08962}.
For NAS, we use AdamW optimizer with a zero weight decay.
We use a batch size of 32 for training and set the learning rate to $5\times 10^{-5}$.


\paragraph{\textbf{Model deployment.}} We deploy our model on different MCUs,
i.e., NUCLEO-F746 with 320KB SRAM and 1MB Flash, NUCLEO-F767 with 512KB SRAM and 2MB Flash as well as NUCLEO-H7A3ZI-Q
with 1.4 MB SRAM and 2MB Flash, to measure the latency and the peak memory
usage. The batch size is fixed to 1. 

\begin{table}[!tb]
    \centering
    \caption{Accuracy comparison on MNLI. The ratio of the first cluster is not shown as it equals to embedding dimension (Emb stands for embedding, params stands for parameters, and Acc stands for accuracy).}
    \label{tab:accuracy_comp}
    \scalebox{0.76}{
    \begin{tabular}{c|cccccc}
    \toprule
    \multirow{2}{*}{Model} & \multirow{2}{*}{Clustering Cutoffs}& \multirow{2}{*}{Ratios}  & Emb  & Total  & \multirow{2}{*}{Acc (\%)} \\ 
    & & &Params (M) &Params (M) &  \\ 
    \midrule
    BERT-tiny      & - & - &3.907M& 4.300M  &  70.01\% \\
    FWSVD          & - & - &0.368M& 0.761M  &  60.83\% \\
    Adaptive Emb & 1000,4000,10000  & 32,8,2 & 0.318M & 0.712M  & 67.01\% \\
    \midrule
    \multirow{2}{*}{MCUBERT-tiny}        &  510,1065, 1915  & 109, 18, 2 & 0.231M &                                       0.624M  &  68.33\% \\
                                    &  1218, 2534, 4061 & 54, 2, 2 & 0.307M & 0.700M  &  69.18\% \\
    \midrule \midrule
    BERT-mini      & - & - & 7.814M& 10.960M & 74.80\%  \\
    Adaptive Emb & 1000,4000,10000  &  64, 16, 4 & 0.648M  & 3.794M & 69.25\% \\
    \midrule
    \multirow{3}{*}{MCUBERT-mini}        &  1014, 3348, 4912 &  41, 41, 2 & 0.492M & 3.638M  & 71.89\% \\
                                    &  1354, 4301, 6094 &41, 42, 2 & 0.613M & 3.759M  & 72.59\%  \\
                                    &  1871, 5591, 7788 &  52, 18, 2 & 0.779M &  3.925M & 74.22\%  \\
    \bottomrule
    \end{tabular}
    }
\end{table}


\begin{figure}[!tb]
    \centering
    \includegraphics[width=0.7\linewidth]{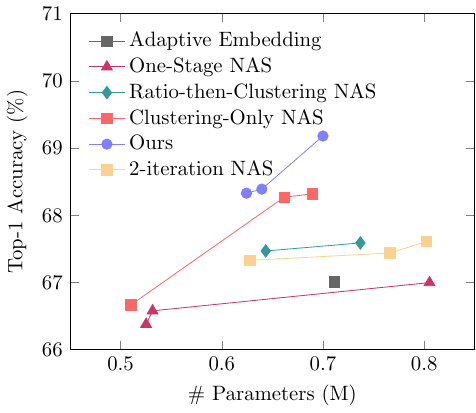}
    \caption{Importance of
    two-stage NAS.}
    \label{fig:importance of two-stage}
    \vspace{-7pt}
\end{figure}

\subsection{Importance of two-stage NAS}
As described in Section \ref{sec:Method}, we propose a two-stage NAS strategy, searching token clustering in the first stage and approximation ratios in the second stage.
To demonstrate the importance of such two-stage formulation,
we compare \method~with other baselines based on BERT-tiny and MNLI dataset, as shown in Figure~\ref{fig:importance of two-stage},
including 1) baseline adaptive embedding;
2) one-stage NAS that searches the token clustering and approximation ratios together (DARTS \cite{liu2018darts});
3) searching approximation ratios first followed by token clustering (denoted as Ratio-then-Clustering NAS);
4) clustering-only NAS that only searches the token clustering;
and 5) 2-round NAS that repeats the two-stage NAS for 2 rounds.
As shown in Figure~\ref{fig:importance of two-stage}, our two-stage NAS outperforms all the other strategies and achieves the best Pareto front.
One-stage NAS that ideally has the largest search space produces the worst Pareto front.
This is because of the large discrepancy among optimal low-rank approximation ratios for different token clustering, which brings convergence difficulties as discussed in Section \ref{sec:Method}.


\begin{table}[!tb]
    \centering
    \caption{Accuracy comparison on other GLUE datasets (Emb stands for 
    embedding, and params stands for parameters).}
    \label{tab:accuracy_comp_other}
    \scalebox{0.9}{
    \vspace{3pt}
    \begin{tabular}{cc|c|c|c}
    \toprule
    Model                     & Metrics & MRPC    &  SST2   & QQP  \\ 
    \midrule
    \multirow{2}{*}{BERT-tiny}& Emb Params  & 3.91M   & 3.91M   &   3.91M    \\
                              & Accuracy     & 74.02\%  & 82.45\%  &  87.16\%   \\
    \midrule
    \multirow{2}{*}{Adaptive Emb} &Emb Params & 0.32M   & 0.32M  &  0.32M   \\
                                  & Accuracy      & 70.34\%  & 81.42\% &  84.32\%   \\
    \midrule
    \multirow{2}{*}{MCUBERT-tiny}      & Emb Params & 0.31M   & 0.26M  &  0.32M   \\
                                  & Accuracy       & 73.77\%  & 82.11\% &  85.20\%   \\
    \midrule \midrule
    Model                     & Metrics & MRPC    &  SST2   & QQP  \\ 
    \midrule
    \multirow{2}{*}{BERT-mini}& Emb Params  & 7.81M   & 7.81M   &   7.81M    \\
                              & Accuracy         & 78.90\%  & 85.32\%  &  89.25\%   \\
    \midrule
    \multirow{2}{*}{Adaptive Emb} &Emb Params & 0.65M   & 0.65M  &  0.65M   \\
                                  & Accuracy       & 75.49\%  & 83.60\% &  87.93\%   \\
    \midrule
    \multirow{2}{*}{MCUBERT-mini}      & Emb Params & 0.65M   & 0.49M  & 0.52M   \\
                                  & Accuracy       & 77.94\%  & 83.83\% &  88.12\%   \\
    \bottomrule
    \end{tabular}
    }
\end{table}

\subsection{Accuracy Comparison on GLUE}

\paragraph{\textbf{Comparison on MNLI.}} We compare \method~with the baseline FWSVD
\cite{hsu2022language} and adaptive embedding \cite{baevski2018adaptive} on the
MNLI dataset, as shown in Table~\ref{tab:accuracy_comp}. 
For both
BERT-tiny and BERT-mini, \method~can simultaneously achieve better accuracy and smaller
parameter size compared to the baseline FWSVD and adaptive embedding. Specifically,
for BERT-tiny, \method~can achieve 1.4$\times$ and 1.6$\times$ embedding parameter reduction
with $1.3$\% and $7.5$\% better accuracy compared to adaptive embedding and FWSVD,
respectively. With the same model size, \method~improves the accuracy by $2.2$\% and $8.4$\%.
For BERT-mini, \method~achieves $3.3$\% better accuracy with a smaller parameter size compared
to adaptive embedding. We also observe for BERT-tiny, \method~
tries to assign a higher approximation ratio for the second cluster
but assign more tokens to the fourth cluster.

\paragraph{\textbf{Comparison on Other Datasets.}}
We then compare MCUBERT with adaptive embedding on other GLUE datasets.
Note we focus on the MRPC, SST2, and QQP datasets as even the full BERT-tiny and BERT-mini
suffers from accuracy issues on COLA and RTE datasets~\cite{devlin2018bert}.
As shown in Table~\ref{tab:accuracy_comp_other}, \method~consistently out-performs
adaptive embedding on all three tasks by 3.43\%, 0.69\%, 0.88\% better accuracy
for BERT-tiny, respectively. For different datasets,
the number of tokens and the approximation ratio
of each cluster are different, indicating the necessity of task loss-guided compression.


\begin{figure*}[!tb]
    \begin{minipage}{0.60\linewidth}
        \centering
        \subfloat[\label{fig:a}]{\includegraphics[width=0.5\linewidth]{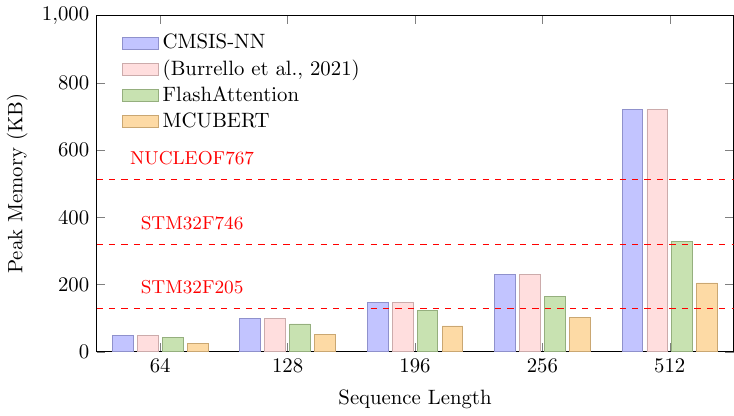}}
        \subfloat[\label{fig:b}]{\includegraphics[width=0.5\linewidth]{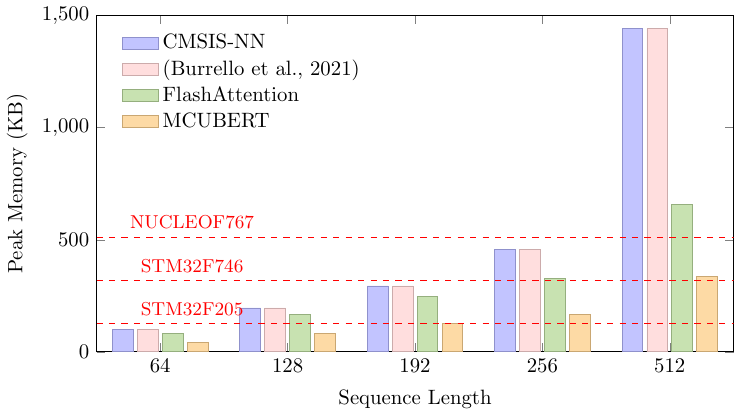}}
        \caption{Peak memory comparison for different sequence lengths for (a) BERT-tiny and (b) BERT-mini.}
        \label{fig:exp:memory-sequence length}
    \end{minipage}
    \hfill
    \begin{minipage}{0.33\linewidth}
        \centering
        \includegraphics[width=\linewidth]{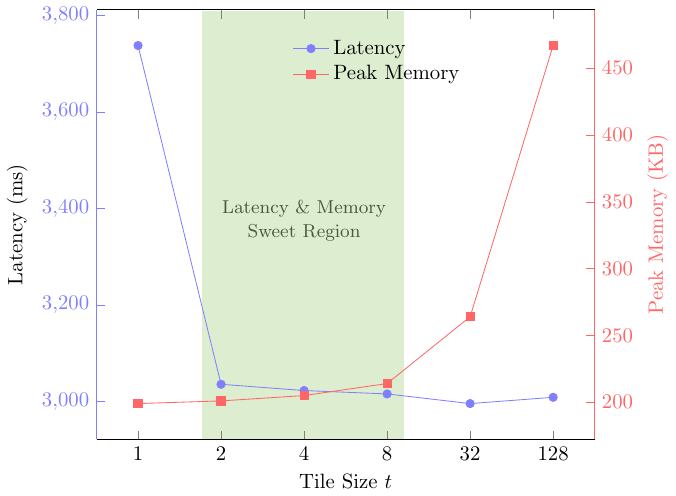}
        \caption{Comparison of latency and memory with different tile sizes $t$.}
        \label{fig:exp:latency memory-t}
    \end{minipage}
\end{figure*}

\begin{figure}[!tb]
    \centering
    \includegraphics[width=0.95\linewidth]{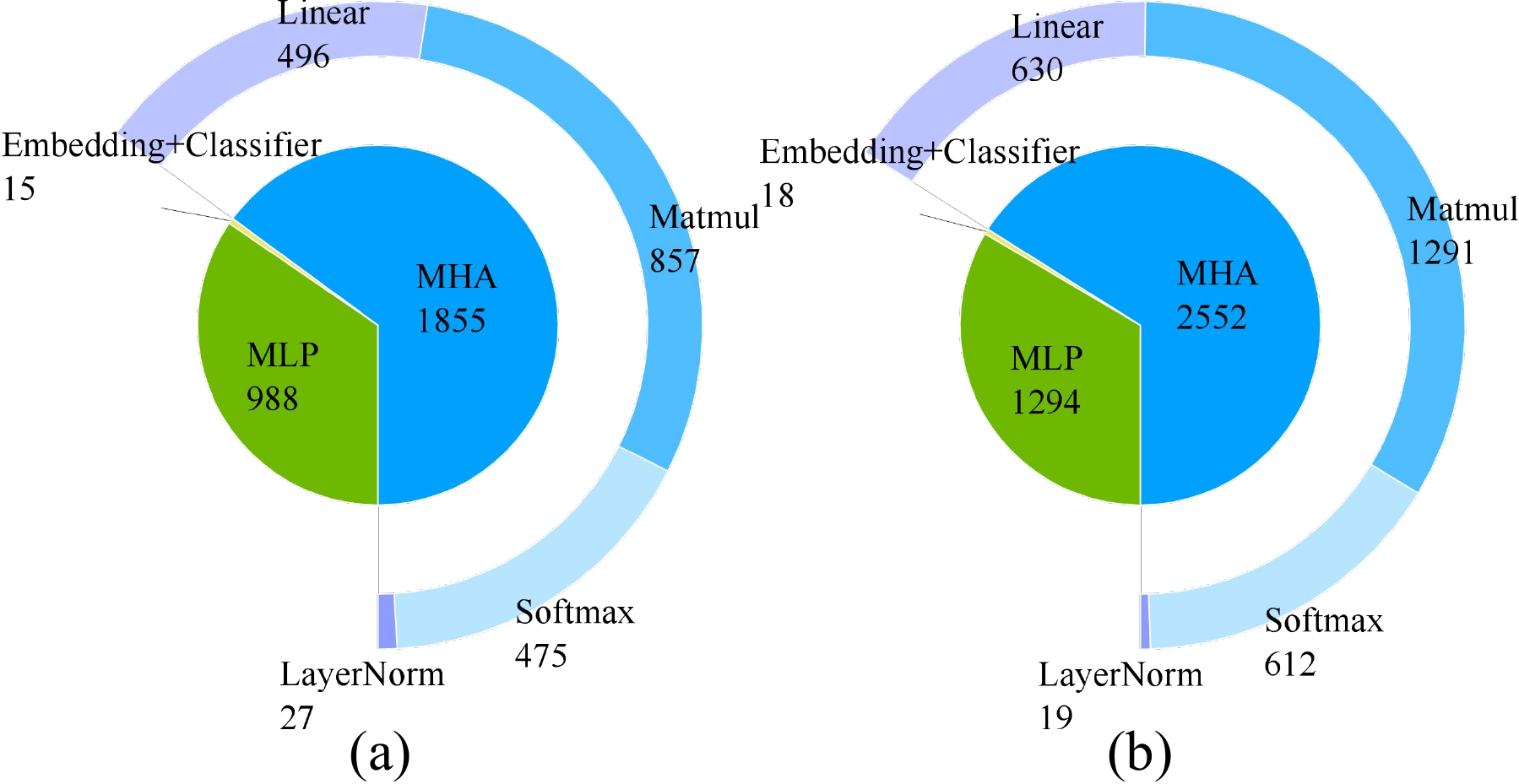}
    \caption{Latency (ms) breakdown for (a) \method~and (b) \cite{burrello2021microcontroller} with 512 sequence length
    (Matmul stands for matrix multiplication.).}
    \label{fig:latency_breakdown}
    \vspace{-7pt}
\end{figure}

\subsection{Peak Memory and Latency Comparison}

\paragraph{\textbf{Peak memory comparison.}} We compare the inference latency and peak memory
with the baseline CMSIS-NN \cite{lai2018cmsis}, \cite{burrello2021microcontroller}, and FlashAttention \cite{dao2022flashattention}.
We re-implement \cite{burrello2021microcontroller} and \cite{dao2022flashattention} based on the original paper. When re-implement FlashAttention, we need to store the output matrix in high precision, which has been discussed in Section \ref{sec:schedule_opt}.
As shown in Figure~\ref{fig:exp:memory-sequence length}, both CMSIS-NN and \cite{burrello2021microcontroller}
do not consider the execution memory in the scheduling, leading to a fast memory growth and out-of-memory 
issue for a long sequence length, i.e., 512. For BERT-tiny, \method~achieves more than 1.9$\times$ and 3.5$\times$
peak memory reduction compared to the baseline when the sequence lengths are 64 and 512, respectively.
This indicates \method~can support 3$\times$, 2$\times$, and 2$\times$ longer
input sequences compared to CMSIS-NN and \cite{burrello2021microcontroller}
on MCUs with 512 KB, 320 KB and 128 KB SRAM, respectively.
In FlashAttention, the output matrix needs iterative updates, demanding fp32 data format instead of int8 precision, causing substantial memory usage growth. Moreover, typical FlashAttention implementation keeps the entire query, key, and value matrices in memory, further increasing memory consumption compared to our approach. Compared with FlashAttention, MCUBERT also achieves more than 1.6$\times$ and 1.9$\times$
peak memory reduction in all sequence lengths for BERT-tiny and BERT-mini, respectively.

\paragraph{\textbf{Latency comparison.} }
The latency comparison for different sequence lengths is shown in Table~\ref{tab:latency_comp_seq}.
\method~achieves around 1.5$\times$ and 1.3$\times$ latency reduction compared to CMSIS-NN and \cite{burrello2021microcontroller} 
consistently for different
sequence lengths and different MCUs. We visualize the latency breakdown for \cite{burrello2021microcontroller} and \method.
As shown in Figure~\ref{fig:latency_breakdown}, \method~reduces the latency of both
the linear and matrix multiplication operators and fuses all tensor shape transformation operators,
demonstrating the high efficiency of our kernel design. Compared with other acceleration strategies such as FlashAttention,  MCUBERT still achieves lower latency, although MCUBERT already shows lower peak memory usage than FlashAttention.
\textit{
With the model and scheduling optimization, \method~enables to deploy BERT-tiny on NUCLEO-F746 with sequence length of 512 for the first time.
}
\begin{table}[!tb]
    \centering
    \caption{Latency (ms) comparison on NUCLEO-F767 (F7) and NUCLEO-H7A3ZI-Q (H7) for different sequence lengths using BERT-tiny (OOM stands for out of memory).}
    \label{tab:latency_comp_seq}
    \scalebox{1.0}{
    \vspace{3pt}
    \begin{tabular}{cc|ccccc}
    \toprule
    MCU                  & Methods                            & 64 & 128 & 192 & 256 & 512  \\ 
    \midrule
    \multirow{4}{*}{F7}  & CMSIS-NN                           & 386 & 838 & 1355 & 1939  &  OOM  \\
                         & \cite{burrello2021microcontroller} & 366 & 780 & 1263  & 1812  &  OOM  \\
                         & FlashAttention                            & 273 & 606 & 880  & 1463  & 4012   \\
                         & \method                            & 264 & 578 & 942  & 1354  & 3591   \\
    \midrule
    \multirow{4}{*}{H7}  & CMSIS-NN                           & 320 & 697 & 1138 & 1638  & 4233   \\
                         & \cite{burrello2021microcontroller} & 284 & 625 & 1016  & 1471  & 3866   \\
                         & FlashAttention                            &220 & 491 & 818  & 1205  &  3297  \\
                         & MCUBERT                            & 213 & 469 & 768  & 1116  & 2940   \\
    \bottomrule
    \end{tabular}
    }
\end{table}

\subsection{Ablation Study}
\begin{figure}[!tb]
    \centering
    \includegraphics[width=0.7\linewidth]{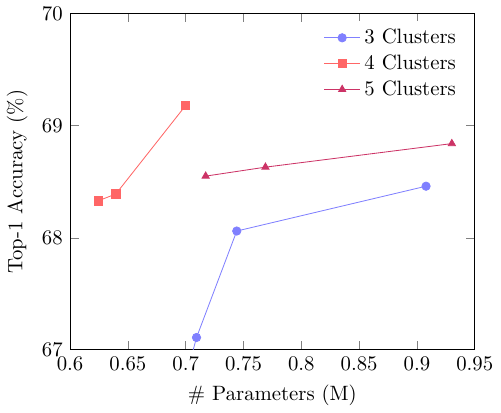}
    \caption{Impact of the number of clusters.}
    \label{fig:exp:impact of number of cluster}
    \vspace{-7pt}
\end{figure}

\paragraph{\textbf{Impact of the number of clusters.}} \method~follows adaptive embedding and empirically fix
the number of clusters to 4. We now evaluate its impact on the
accuracy-parameter Pareto front. We
use the BERT-tiny model and MNLI dataset for comparison.
As shown in Figure~\ref{fig:exp:impact of number of cluster}, the Pareto
front for $c = 4$ indeed outperforms that for $c=3$ or $c=5$.  
We observe $c=3$ performs much worse when the model parameter size is small.
We hypothesize this is because when $c=3$, important tokens are forced to have a small dimension, leading to low accuracy.

\paragraph{\textbf{Impact of fine-grained scheduling.}}
To evaluate the impact of fine-grained scheduling on both latency and memory,
we change the number of tokens in each tile, i.e., $t$.
The change in latency and peak memory is plotted in Figure~\ref{fig:exp:latency memory-t}. 
As we can observe, the peak execution memory reduces consistently with the decrease of $t$ while
the latency remains roughly the same for $t \leq 2$. 
The high latency for $t=1$ is because of hardware under-utilization.

\section{Conclusion}
This work proposes MCUBERT, a network/scheduling co-optimization framework enabling BERT on MCUs.
For network optimization, MCUBERT proposes an MCU-aware two-stage NAS algorithm with clustered low-rank approximation for embedding compression.
For scheduling optimization, we leverage tiling, in-place computation, and kernel optimization to simultaneously reduce peak memory and latency.
\method~overcomes all existing baselines and enables to run the lightweight BERT models on commodity MCUs for the first time.

\begin{acks}
This work was supported in part by the NSFC (62125401), the 111 Project (B18001), the Theme-Based Research
Scheme (TRS) Project T45-701/22-R and General Research Fund (GRF) Project 17203224 of the Research Grants Council (RGC), HKSAR.
\end{acks}
\newpage
\bibliographystyle{ACM-Reference-Format}
\bibliography{acmart}

\end{document}